\pdfoutput=1

\documentclass[11pt]{article}

\usepackage{acl}

\usepackage{times}
\usepackage{latexsym}
\usepackage{multirow}
\usepackage{booktabs} 
\usepackage{tabularx} 
\usepackage{tabulary} 
\usepackage{float} 
\restylefloat{table}
\usepackage{siunitx} 

\usepackage[T1]{fontenc}

\usepackage[utf8]{inputenc}

\usepackage{microtype}

\newcolumntype{Z}{>{\raggedleft\let\newline\\\arraybackslash\hspace{0pt}}X}

\usepackage{tcolorbox}
\usepackage{pythonhighlight}

\title{Evaluating Large Language Models for Document-grounded Response Generation in Information-Seeking Dialogues}

\author{Norbert Braunschweiler \and Rama Doddipatla \and Simon Keizer \and \\
{\bf Svetlana Stoyanchev} \\
 Toshiba Europe Limited \\
 Cambridge Research Laboratory, 208 Cambridge Science Park \\
 Cambridge, UK \\
 \texttt{\{norbert.braunschweiler,rama.doddipatla,simon.keizer,} \\
 \texttt{svetlana.stoyanchev\}@toshiba.eu} \\}

\begin{document}
\maketitle



\begin{abstract}

In this paper, we investigate the use of large language models (LLMs) like \emph{ChatGPT}
for document-grounded response generation in the context of information-seeking dialogues. 
For evaluation, we use \emph{MultiDoc2Dial} corpus of task-oriented dialogues in four social service domains
previously used in the \emph{DialDoc 2022 Shared Task}. 
Information-seeking dialogue turns are grounded in multiple documents providing relevant information.
We generate dialogue completion responses by prompting a \emph{ChatGPT} model, using two methods:
\emph{ChatCompletion} and \emph{LlamaIndex}.
\emph{ChatCompletion} uses knowledge from \emph{ChatGPT} model pre-training while \emph{LlamaIndex}
also extracts relevant information from documents.
Observing that document-grounded response generation via LLMs cannot be adequately assessed
by automatic evaluation metrics as they are significantly more verbose, we perform a human evaluation
where annotators rate the output of the shared task winning system,
the two \emph{ChatGPT} variants outputs, and human responses. 
While both \emph{ChatGPT} variants are more likely to include information not present in the relevant segments,
possibly including a presence of hallucinations, they are rated higher than
both the shared task winning system and human responses.

\end{abstract}



\section{Introduction}

Accessing domain-specific knowledge in task-oriented dialogue modeling is a crucial aspect to
provide information-seeking users with relevant, trustworthy and detailed information. Often this knowledge
has to be retrieved from various knowledge sources stored in diverse formats and multiple documents.
Once the relevant knowledge has been retrieved, a dialogue system then needs to combine it with the dialogue context and user query 
to generate an informed, coherent and fluent natural language response.

In this paper, we present a comparison of methods for knowledge-grounded response generation in
task-oriented dialogues which include traditional retrieval-augmented generation models as well
as state-of-the-art large language models (LLMs), such as \emph{ChatGPT}~\cite{Ouyang_InstructGPT_2022},
one of the \emph{GPT-model} variants released by \emph{OpenAI}~\cite{ChatGPT_2023}.
While there is a wide range of possibilities to utilize LLMs for this task, we focus on two 
methods of prompting LLMs to investigate their capabilities for this
particular scenario. 

The first method, uses a chat-interface (``ChatCompletion''\footnote{\url{https://platform.openai.com/docs/guides/chat/chat-completions-beta}})
which takes as input 1) a dialogue context, 2) a short description of the system role
and its domain, and 3) the last user utterance. It thus effectively provides the LLM with context information
about the topics in conversation so far and the general topic the user query is rooted in. 
The second method uses the \emph{LlamaIndex}~\cite{Liu_LlamaIndex_2022} tool
that combines information extraction from multiple documents with the LLM. 
In this approach, dialogue context is used to extract relevant information from indexed documents
and passed as context to the LLM to generate a grounded response.

Our study assesses how suitable these two methods are for the given task by using 
the \emph{MultiDoc2Dial}~\cite{Feng_2021_MultiDoc2Dial} corpus for evaluation
which includes task-oriented dialogues grounded in multiple documents. 

Our paper is structured as follows: Following the description of  the \emph{MultiDoc2Dial} corpus
in the next section, the  \emph{DialDoc Shared Task} is introduced which includes the test-set utilized in this paper.
Then, the four response generation methods compared in our study are described in detail, in particular
focusing on the two new methods accessing the \emph{GPT-model}. This is followed by a definition of the experimental design
and a presentation of the results of objective and human evaluations and finished in a conclusion.


\section{MultiDoc2Dial corpus}
\label{sec:MultiDoc2Dial_corpus}

The \emph{MultiDoc2Dial} dataset\footnote{\url{http://doc2dial.github.io/multidoc2dial/}}
contains 4.8k dialogues (61078 turns) between an information-seeking user and an agent.
Dialogues include an average of 14 turns and are grounded in 488 documents from 4 domains:
Department of Motor Vehicles (\emph{dmv}), Social Security Administration (\emph{ssa}),
Federal Student Aid (\emph{sta}) and Veterans Affairs (\emph{va}).
Documents include HTML mark-ups (e.g.\ title, list) and document section information (title, text body, spans/sections).
The dialogue data includes annotations at the turn-level for dialogue act, speaker role, human-generated utterance and
the associated grounding span with document information.
We chose this dataset for our evaluation as it provides
a) task-oriented dialogues in which agent responses are grounded in multiple documents from four domains,
b) the content of these documents, and
b) manually entered agent responses plus gold standard passages from associated documents 
which provide relevant grounding information extracted from the document that
may be used to assess the correctness of model generated responses.

Table~\ref{tab:MultiDoc2Dial_corpus} shows an overview of the \emph{MultiDoc2Dial}~\cite{Feng_2021_MultiDoc2Dial} corpus
providing the number of documents and dialogues in each domain as well as the number of dialogues which include one or
more segments where a segment includes turns grounded in the same document.

\begin{table}[t]
\center
\small
\begin{tabular}{l|ccccc}
Domain & \#doc & \#dial & 2seg & >2seg & single\\
\toprule
ssa       & 109    & 1191   & 701      & 188       & 302 \\
va        & 138    & 1337   & 648      & 491       & 198 \\
dmv     & 149    & 1328   & 781      & 257       & 290 \\
sta       & 92     & 940     & 508      & 274       & 158 \\
\midrule
total     & 488    & 4796   & 2638    & 1210      & 948 \\
\bottomrule
\end{tabular}
\caption{Statistics of the \emph{MultiDoc2Dial} corpus.}
\label{tab:MultiDoc2Dial_corpus}
\end{table}


\subsection{DialDoc Shared Task}
The \emph{MultiDoc2Dial} corpus was also used as training, validation and test set for the
\emph{DialDoc 2022 Shared Task}\footnote{\url{https://doc2dial.github.io/workshop2022/}}
on ``Open-Book Document-grounded Dialogue Modeling''~\cite{feng-etal-2022-dialdoc}.
This shared task mainly focused on building open-book goal-oriented
information seeking conversation systems, where an agent could provide an answer or
ask follow-up questions for clarification or verification~\cite{feng-etal-2022-dialdoc}.
The main goal was to generate grounded agent responses in natural language based
on the dialogue context and domain knowledge in the documents.
Specifically, taking as input 1) latest user turn, 2) dialogue history and 3) all domain documents and
then generate as output the agent response in natural language.
A sub-task was grounding span prediction which aimed at locating related spans from
multiple documents. A summary of the shared task is provided in~\citealp{feng-etal-2022-dialdoc}.

Table~\ref{tab:MultiDoc2Dial_corpus_splits} shows the splits of the \emph{MultiDoc2Dial} corpus
into sets for training, validation and testing used in the \emph{DialDoc 2022 Shared Task}. 

\begin{table}[t]
\center
\small
\begin{tabular}{l|ccc}
                            & Train   & Val   & Test\\
\toprule
\#dialogues            & 3474   & 661  & 661\\
\#queries               & 21453 & 4201 & 4094\\
avgQueryLength      & 104.6 & 104.2 & 96.5\\
avgResponseLength & 22.8   & 21.6  & 22.3\\
\bottomrule
\end{tabular}
\caption{Train/val/test-splits of \emph{MultiDoc2Dial}.}
\label{tab:MultiDoc2Dial_corpus_splits}
\end{table}

As part of the \emph{DialDoc Shared Task}, organizers provided a baseline model and published
a leaderboard\footnote{\url{https://eval.ai/web/challenges/challenge-page/1437/leaderboard/3575}} of participants models performances
in various stages during the shared task.  For our comparison, we are using both this baseline model as well as the responses generated by the
shared task winning team CPII-NLP~\cite{li-etal-2022-grounded}, in the \emph{SEEN} evaluation setting
where the test dialogues shared the same domains as the training data~\cite{feng-etal-2022-dialdoc}. 

The shared task evaluation is based on a subset of 661 turns selected from the 4094 test-set turns. These
661 turns also formed the test-set of the models compared in this study, including the
 GPT-based models introduced in the next section.



\section{Response generation methods}
\label{sec:KnowledgeGroundedResponseGeneration}

This section introduces the response generation methods compared in this paper. 
The selection of methods was based on approaches which could a) serve as baselines because they
had published performance results on the \emph{MultiDoc2Dial} corpus, b) were available at the time of writing, 
c) utilized the capabilities of the \emph{GPT-models}, and d) could be evaluated in both objective and
human evaluation. Since the character of this study is to get an initial indication of the abilities of \emph{GPT-models} for
the given task of document-grounded response generation for information-seeking dialogue modeling, 
we restricted the number of approaches to a total of four, including two baseline models using established
architectures and two methods employing \emph{GPT-models}.  

\subsection{Baselines}
\label{ssec:Baselines}

As baselines, we selected two of the models which were part of the
\emph{DialDoc 2022 Shared Task}~\cite{feng-etal-2022-dialdoc} described in the previous section~\ref{sec:MultiDoc2Dial_corpus},
i.e., the baseline model for the \emph{DialDoc Shared Task} (henceforth called \emph{RAGBase}) and the shared task winning model
from team \emph{CPII-NLP}~\cite{li-etal-2022-grounded} (henceforth called \emph{CPII-NLP}).

The \emph{RAGBase} model applies a Retrieval-Augmented Generation (RAG)~\cite{lewis2021retrievalaugmented} architecture
described in~\citealp{Feng_2021_MultiDoc2Dial} which combines a retriever model employing a
\emph{Dense Passage Retriever (DPR)}~\cite{karpukhin-etal-2020-dense} with a generator adopting
the \emph{BART-large}~\cite{lewis-etal-2020-bart} model which was 
pre-trained on the CNN dataset~\cite{Feng_2021_MultiDoc2Dial}.

Model \emph{CPII-NLP}, which significantly outperformed the baseline model in the \emph{DialDoc Shared Task}, 
also includes a \emph{Dense Passage Retriever (DPR)} and a \emph{BART-large} model for generation, but extends this
architecture with a re-ranker (following the retrieval step) that includes an ensemble of 3 cross-encoder models using
\emph{BERT}~\cite{devlin2019bert}, \emph{RoBERTa}~\cite{Liu2019RoBERTaAR},
\emph{ELECTRA}~\cite{Clark2020_ELECTRA}, while the \emph{BART-large} model was jointly trained
with a grounding span predictor.
The 3 components are individually optimized, while passage dropout and regularization techniques are adopted to improve
the response generation performance.

The two baseline models provided benchmarks for assessing the two new models
introduced next, which use two variants of utilizing one \emph{GPT-model} for knowledge-grounded response generation.

\subsection{GPT-based models}
\label{ssec:GPT_based_Models}

To enable an equal comparison we selected one LLM from the \emph{GPT-models} repertoire
available from OpenAI\footnote{\url{https://openai.com/product\#made-for-developers}}:
The \emph{gpt-3.5-turbo} model. It includes optimizations for chat
and usage costs are significantly lower (1/10\textsuperscript{th}) than for the \emph{text-davinci-003}
model\footnote{\url{https://platform.openai.com/docs/models/gpt-3-5}}.

The \emph{gpt-3.5-turbo} model 
has a token input limit of 4096 tokens, uses training data up to September 2021
and has usage costs of \$0.002/1k tokens.

We select two variants of accessing \emph{gpt-3.5-turbo} motivated by 
their availability at the time of writing
and their capabilities to a) represent dialogue context (\emph{GPTChat})
and b) retrieving relevant knowledge from the associated documents (\emph{GPTLama}). 

The goal of this comparison is to evaluate these models in the selected scenario of knowledge-grounded response generation for
task-oriented dialogue introduced in section~\ref{sec:MultiDoc2Dial_corpus}.
Generating a response for a user query which takes a) dialogue context into account,
and b) knowledge retrieved from multiple documents. 

Table~\ref{tab:Methods4knowledgeGroundedResponseGeneration} shows the four systems compared in this study. 
Next, the two models for knowledge-grounded response generation utilizing the \emph{gpt-3.5-turbo} LLM
are introduced in detail.

\begin{table}[t]
\center
\begin{tabularx}{\linewidth}{l|X}
{\bf Method}        & Features\\
\toprule
\emph{RAGBase}  & Retrieval-Augmented Generation (RAG), DialDoc baseline model\\
\emph{CPII-NLP}  & Pipeline system of retriever, re-ranker, generator, DialDoc winner\\
\emph{GPTChat}  & GPT-ChatCompletion-API \& system intro prompt, no grounding\\
\emph{GPTLama} & GPT \& Knowledge-grounded prompt generation via LlamaIndex\\
\bottomrule
\end{tabularx}
\caption{Methods for knowledge-grounded response generation which are compared in this paper.}
\label{tab:Methods4knowledgeGroundedResponseGeneration}
\end{table}



\subsubsection{ChatCompletion method: \emph{GPTChat}}
\label{ssec:ChatCompletion method}

OpenAI provides a \emph{ChatCompletion}-module\footnote{\url{https://platform.openai.com/docs/guides/chat/chat-completions-beta}}, which
allows input to \emph{GPT-models} in the form of a structured dialogue including user-queries and agent-responses. 
The model can then use the dialogue context to generate responses in the agent-role for a given user query.
While this method does not provide a knowledge retrieval step, it still can be guided by providing additional
``system''-messages which can include instructions such as {\it You are a helpful assistant} or background knowledge
such as the domain the user query refers to,
e.g.\ {\it Hello, this is the service agent from the U.S. Department of Motor Vehicles – how can I help you?}.

The \emph{GPTChat} system, therefore, provides a reference for how well the \emph{GPT-model} 
performs on the given task without having additional knowledge extracted from the associated documents, but instead
relies on the capabilities of  the \emph{GPT-model} to understand the input and retrieve information from the data it was trained on.
As the topics covered in the four domains of the \emph{MultiDoc2Dial} corpus were in the public domain (websites),
it is likely that they were part of the \emph{GPT-model's} training data. 
But even if it was part of the training data, there is still the challenge of understanding the user input in the context of the dialogue
and then formulating an adequate answer for it. 

The dialogues from the test-set of the \emph{MultiDoc2Dial} corpus were used as input to
the \emph{ChatCompletion-API}\footnote{\url{https://platform.openai.com/docs/guides/chat/introduction}}
by providing both user questions and agent-responses
in the chat format required by the API and shown in Table~\ref{tab:ChatCompletionInptFormatExample}
in appendix~\ref{GPTChat_Exampleofinput}.

To provide the \emph{GPT-model} some context information about the domain of the conversation
it was given an initial system-prompt with a domain-specific content in the ``system'' field; these prompts are listed in
section~\ref{ssec:InitialSysPromptsChatCompletion} in the appendix, and 
were intended to specify the role of the \emph{assistant} and its behavior. 

Each dialogue in the test-set was first split into sub-dialogues at the turn level, always ending with a user turn
in order to get an ``assistant''-response generated by the \emph{GPT-model}. As such, each split formed an individual
dialogue for the \emph{GPT-model} with its own context including all previous turns up to the user turn to be answered until
the full dialogue was represented. 
Therefore, the amount of context was growing for each additional turn in the dialogue. 
However, none of the dialogues in the test-set exceeded the token limit of 4096 set in the \emph{gpt-3.5-turbo} model. The
average number of tokens in the full dialogues was 413 with a maximum of 669 tokens.



\subsubsection{LlamaIndex method: \emph{GPTLama}}
\label{ssec:GPTLama}

In contrast to the \emph{GPTChat} model, which does not access 
the associated documents of the
\emph{MultiDoc2Dial} corpus for grounding its responses, another method was selected
which provided that functionality: \emph{LLamaIndex}~\cite{Liu_LlamaIndex_2022}.
This tool, which we used in our system henceforth called \emph{GPTLama},
enables the connection of LLM's with external or private data. It provides tools to load external/private
documents and parse them into data structures which can be queried efficiently
to retrieve relevant information for a given user input both of which 
can then be combined into a prompt send to a chosen LLM.

By design, the quality of the response will therefore largely depend on the data retrieved by the query of the indexed documents
and to a lesser extend on the language reasoning capabilities of the LLM.

By using \emph{LlamaIndex} tools\footnote{We used LlamaIndex version 0.6.0} we aimed at improved accuracy of response generation by grounding
it in the associated documents. 
Since it first retrieves knowledge from documents and then sends it together with the user input to the LLM
for response generation its closer to the
retrieval-augmented response generation method in the two baseline models.  

We used \emph{semantic search}\footnote{\url{https://gpt-index.readthedocs.io/en/latest/use_cases/queries.html\#semantic-search}} for queries
over our documents which were first converted into a \emph{LlamaIndex} vector store by saving each of the 488 documents into a single file
and running the {\it GPTVectorStoreIndex.from\_documents(documents)}\footnote{\url{https://gpt-index.readthedocs.io/en/latest/getting_started/starter_example.html\#build-and-query-index}} command across these individual files. 

One of the challenges in prompt creation for \emph{LlamaIndex} is to ensure that the user input provides sufficient context information for
a) querying the vector index and b) using it for the response generation from the \emph{GPT-model}. In our scenario of document-grounded 
response generation for task-oriented dialogue modeling, an isolated user turn, i.e., the last utterance of dialogue context, 
might not include sufficient content to retrieve relevant information
from the associated documents. In \emph{MultiDoc2Dial} corpus, a typical initial user input asks a specific question related to one of the four domains and
includes already information to link it to one of these domains. However, some inputs can be as simple as {\it Hi there} or
{\it I need to know how to apply, please} which do not indicate any particular domain and are therefore not useful for retrieving relevant
information from the documents. Therefore, to provide sufficient information about the domain of the user input it was provided
as part of the query string, e.g.\ {\it Question for the U.S. department for Veterans Affairs (VA) service agent: } 
(see section~\ref{ssec:InitialSysPrompts4GPTLama} for a full list of prompts)
and then followed by the user query. This way the retrieval step had at least a chance to locate domain related documents. 
In addition, by specifying the role of the ``responder'' as ``service agent'' it provided additional instructions for the LLM
how to respond. 

Another aspect is the representation of dialogue context within the \emph{LlamaIndex} framework. Since it does not provide the same
dialogue representation format as in the \emph{ChatCompletion-API} introduced in section~\ref{ssec:ChatCompletion method}
it is not as straightforward to be included. 
To provide dialogue context we used the 
\emph{QuestionAnswer}-prompt\footnote{\url{https://gpt-index.readthedocs.io/en/v0.6.0/how_to/customization/custom_prompts.html\#example}}
method in \emph{LlamaIndex}  which requires both a \emph{query\_str} field and a \emph{context\_str} field, with
the user query included in the \emph{query\_str} field and the \emph{context\_str} including the information retrieved from associated
documents by using the \emph{query\_str} as search query. 

We used the \emph{query\_str} field to enter the last user turn of the dialogue under consideration and loaded the 
\emph{context\_str} with previous dialogue context as well as ``system''-instructions specifying the response behavior of the LLM.
Examples of domain-specific entries used for \emph{query\_str} and enriched \emph{context\_str} are in
appendix sections~\ref{ssec:InitialSysPrompts4GPTLama} and  \ref{ssec:PreDialContextPrompts4GPTLama} respectively.

Because of the additional information from the retrieval step, the \emph{LlamaIndex} method uses more tokens per dialogue call than the
\emph{ChatCompletion} method, while still remaining below the token limit of 4096 for the test-set dialogues.
The average number of tokens send to the \emph{GPT-model} via the \emph{GPTLama} method was 1114 with a maximum of 1590 tokens.


\section{Experimental design}

We compare four response generation methods in a task-oriented dialogue scenario
of which three are using document-grounded generation,
in which user input requires the system to find and retrieve relevant knowledge from multiple documents
and one method relying on the abilities of the selected LLM to retrieve relevant knowledge from its
own training data. 

Methods are compared using a) objective metrics, i.e., \emph{RougeL}~\cite{lin-2004-rouge}, 
\emph{METEOR}~\cite{banerjee-lavie-2005-meteor}, 
token-level \emph{F1-score}~\cite{rajpurkar-etal-2016-squad}, and \emph{SacreBLEU}~\cite{post-2018-call},
and b) by human evaluation. 

\section{Results}


\subsection{Objective evaluation}

\begin{table*}[ht]
\center
\begin{tabular}{l|cccccc}
System    & Avg\#words &  F1    & SacreBLEU & METEOR & RougeL & Total\\
\midrule
\emph{RAGBase}  & $19.2$ $(10.2)$  & $35.59$ & $22.49$ & $34.62$ & $33.84$ & $126.55$\\
\emph{CPII-NLP}  & $20.6$ $(11.4)$  & $52.17$ & $37.46$ & $51.71$ & $50.15$ & $191.51$\\
\midrule
\emph{GPTChat}  & $69.6$ $(20.4)$  & $17.59$ & $2.30$ & $24.18$ & $13.88$ & $57.95$\\
\emph{GPTLama} & $59.2$ $(46.7)$  & $17.33$ & $4.89$ & $22.71$ & $15.30$ & $60.25$\\
\bottomrule
\end{tabular}
\caption{Comparison of response generation methods on \emph{MultiDoc2Dial}-corpus 661-turn test-set.}
\label{tab:results_objective_metrics_method_comparisons}
\end{table*}

Table~\ref{tab:results_objective_metrics_method_comparisons} shows results of objective metrics for the four
knowledge retrieval methods on the 661-turn test-set from the \emph{MultiDoc2Dial} corpus.
We report the locally replicated performance figures in the \emph{SEEN} category for models
\emph{RAGBase} based on the re-implementation of the \emph{DialDoc}-baseline model
and for model \emph{CPII-NLP} based on responses provided by the authors~\cite{li-etal-2022-grounded}.
Both of these performance figures are deviating only marginally from published results. 

It can be seen, that the objective metrics for the \emph{GPT}-based models are significantly lower than both the baseline model as well as the
winning model. One of the reasons for this lower performance is that \emph{GPT}-generated responses are on average
much longer than the relatively short responses in the reference corpus as seen in the average number of words per response column (Avg\#words).
While these objective metrics provide a useful indication for the quality of the response, they are based on the word overlap
between reference and prediction.
As only one `gold' reference is available, the automatic measures may fail to recognize an appropriately
generated response which addresses a different aspect than the 'gold' reference.  

Therefore, we decided to conduct a human evaluation of agent's responses, which is described next.

\subsection{Human evaluation}

\begin{table}[ht]
\center
\begin{tabular}{l|cc}
System                &  Appropriateness & {\small InfoNotInGround}\\
& Q1 & Q3 \\
\toprule
\emph{Reference}  &   $4.07$ $(1.33)$     &  $10.6\%$ \\
\emph{CPII-NLP}   &    $3.90$ $(1.39)$     &  $16.4\%$ \\
\emph{GPTChat}   &    $4.17$ $(1.22)$     &  $88.0\%$   \\
\emph{GPTLama}  &    $4.19$ $(1.26)$     &  $84.0\%$ \\
\bottomrule
\end{tabular}
\caption{Results of human evaluation of response generation methods for appropriateness (5-point Likert scale, 5=completely appropriate) and percent responses containing information that is not in the reference (InfoNotInGround).}
\label{tab:results_human_eval_comparisons}
\end{table}

For the human evaluation, 25 dialogue snippets with the length between three and nine turns
(to reduce cognitive load on the annotators) were randomly selected from the test subset
of the \emph{MultiDoc2Dial} corpus. Each dialogue snippet starts from the beginning of the dialogue
and ends on a user's turn. We refer to these snippets as dialogue context.

As we address a `grounded' dialogue task, each dialogue context is associated with
 \emph{grounding segment(s)} from the relevant documents that guides the agent's response.
 We use four experimental conditions to generate the agent response for each dialogue context:
 \emph{GPTChat}, \emph{GPTLama}, \emph{CPII-NLP}\footnote{The authors kindly shared the generated responses with us},
 and the `gold'  human agent's response (\emph{Reference}).
We include the \emph{Reference} condition in the human evaluation for the comparison
with the automatically generated ratings and  to determine how often the agents actually used
the {\it grounding  segment} to produce their response. 
We omit the baseline from the human experiment as it has already been shown that \emph{CPII-NLP}
significantly outperforms the baseline in human rating~\cite{feng-etal-2022-dialdoc}.

\begin{table*}[ht]
\center
\begin{tabularx}{\linewidth}{l|X|X}
&Question  & Answering options\\
\toprule
Q1& Rate the appropriateness of the last agent's response \emph{(Appropriateness) }     &   1: Completely Inappropriate; 2: Somewhat Inappropriate; 3: Uncertain; 4: Somewhat Appropriate; 5: Very Appropriate\\
\midrule
Q2 & Does the last agent's utterance contain all/some/contradictory RELEVANT information from the reference? \emph{ (InfoInGrounding)} &  `All Relefant info', `Some relevant info', `Contradicts', `None of the above' \\
\midrule
Q3&  Does the last agent utterance contain information that is NOT in the grounding segment?  \emph{(InfoNotInGrounding)} &  yes, no \\
\bottomrule
\end{tabularx}
\caption{Human evaluation questionnaire.}
\label{tab:eval_questions}
\end{table*}

The annotator is presented with a dialogue context, the corresponding grounding segments,
the agent's response from of the experimental conditions, and three questions (see Table~\ref{tab:eval_questions}).
Q1 asks to rank appropriateness of the response in context of the dialogue on a 1-5 scale.
Q2 asks whether the generated response included {\it relevant} information
from the grounding segments\footnote{We noticed that not all information in grounding segments
is relevant and emphasized that the annotators should look only for information in the grounding segments
that is relevant to dialogue context.}. Annotators could select between four options,
including {\it `None of the above'} which may happen when the reference did not contain relevant information.
Q3 aimed to detect whether the agent response contained information other than the reference.
After submitting the response, the annotator could choose to continue to another question.


Table~\ref{tab:results_human_eval_comparisons} shows the results of Q1 (Appropriateness)  and Q3 (InfoNotInGrounding). 
The two \emph{GPT-model} based methods outperform both \emph{Reference} and \emph{CPII-NLP} model.
\emph{GPT-model} responses contain more information, which may have had a positive effect on the perception of the annotators. 
Both \emph{GPT}-based methods show very similar appropriateness scores indicating that the \emph{GPTChat} method providing dialogue context
with an introductory system prompt was sufficient to provide a response judged the most appropriate by the annotators.

As expected for Q3, \emph{Reference} has the lowest proportion of responses with information outside of grounding segments ($10.6\%$). 
Both \emph{GPT-models} used significantly more information than what was  provided in the grounding reference,
with $88.0\%$ for  the \emph{GPTChat} method and a slightly lower percentage for the \emph{GPTLama} system  ($84.0\%$),
which  indicates that the information retrieved from documents seemed to have a small impact on the generated content.

  \begin{figure}[H]
\center
\includegraphics[scale=.58]{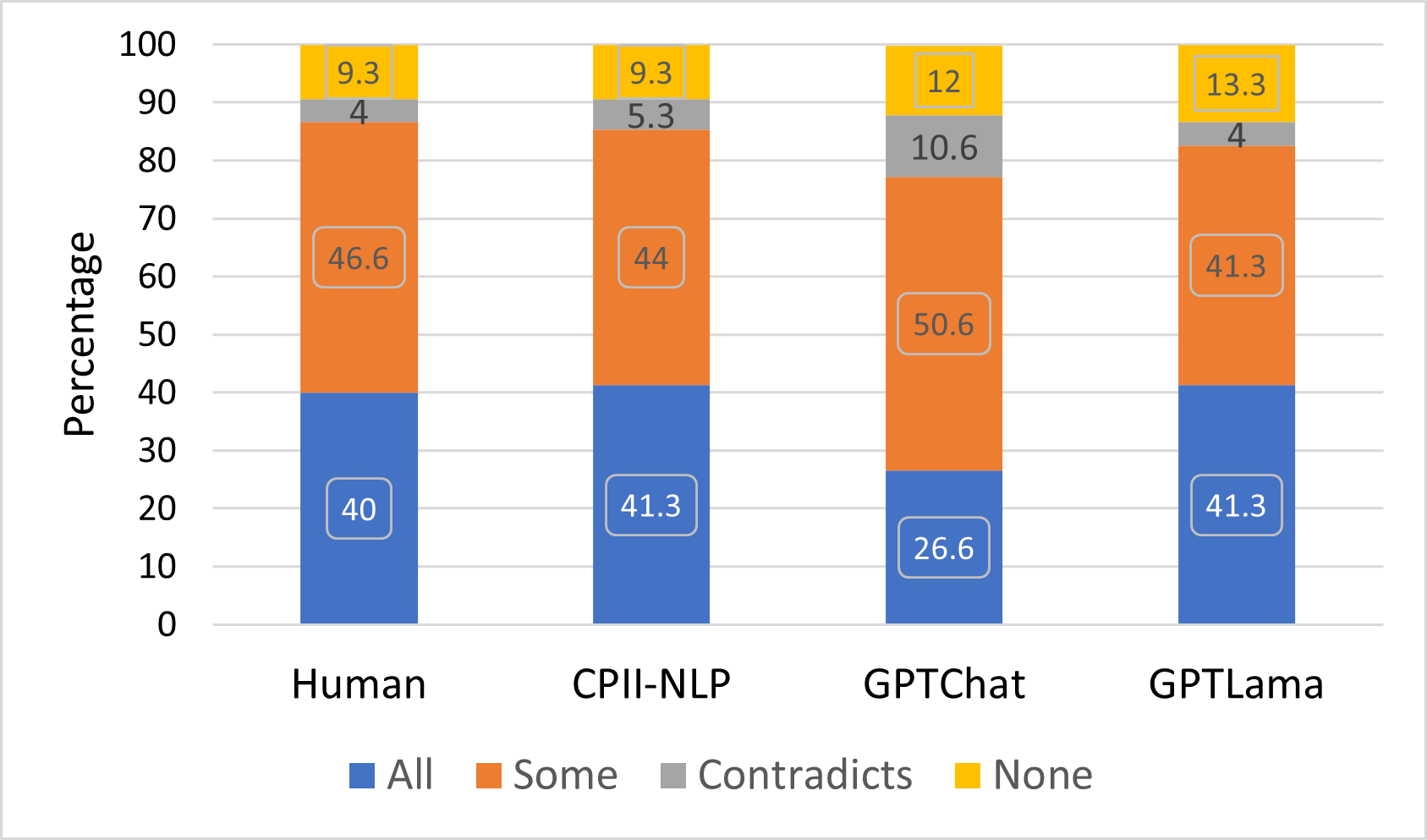}
\caption{Percentage of agent responses containing all/some/contradictory/no information from reference grounding passage (Q2).}
\label{fig:BarPlot-PercentageGoldRefs}
\end{figure}

Figure~\ref{fig:BarPlot-PercentageGoldRefs} shows the percentages of agent responses containing 
all/some/contradictory/no information from the reference grounding passages. 
Results show that human reference as well as system \emph{CPII-NLP} have the highest proportion of responses using {\it all} or {\it some}
information from the grounding segment, followed by system \emph{GPTLama} and the lowest percentage by \emph{GPTChat}.
\emph{GPTChat} has the lowest proportion (26.6\%) with {\it all} and the highest with  {\it some} information (50.6\%).
`None of the above' response for Q2  indicates that a response does not contain relevant information
of the grounding segment or that the grounding segment does not contain any relevant information. 
\emph{GPTChat}  and  \emph{GPTLama} have 12.0\% and 13.3\% of such responses.
Surprisingly, \emph{Reference} also contains 9.3\%  of such responses indicating that grounding segments may not be always relevant.
\emph{GPTChat} has with 10.6\% the highest proportion of responses with contradictory information possible indicating some hallucinations, 
while the remaining three systems range between 4\% and 5.3\% responses ranked as contradicting.

We compute Krippendorff's ${\alpha}$~\cite{Krippendorff_alpha_2008}  to measure the inter-annotator
agreement\footnote{computed between the three annotation instances}. 
The agreement on \emph{Appropriateness} is only  $0.35$. In this subjective task, the annotators may have had different
personal criteria for the agent response appropriateness judgment.
Agreement on \emph{InfoInGrounding} is $0.46$. 
Judging information in grounding segments was open to the annotators' interpretation.
Agreement on \emph{InfoNotInGround} is the highest with ${\alpha}=0.69$ indicating
that the annotators were generally in agreement
whether information outside of the grounding segment was presented.

Examples of model generated responses for a given user question are shown in
Table~\ref{tab:ExamplesOfGeneratedResponses}. 
The example shows that none of the models predicts the correct response of
{\it the civil fine is \$ 200} but models differ in the amount of the fine and while the two baseline models are very close in their wording to the reference only
differing in the amount of the fine, the GPT-generated responses are much longer and either incorrectly retrieve or hallucinate the amount of the fine, but 
formulate their responses very sovereign and convinced.

\begin{table}[H]
\center
\footnotesize
\begin{tabularx}{\linewidth}{lX}
\toprule
           & {\bf Dialogue context}\\
Agent: & Hi, I'm the service agent from the U.S. Department of Motor Vehicles and can help you with any questions related to vehicle registration, drivers licenses or driving fines - how can I help you?\\
User:   & "Hi, I  need to pay a civil penalty. How much is the fine if the order was issued between July 1, 1990 and October 31, 1996?"\\
          & Relevant information from documents: {\it "if the order was issued between January 1 , 1981 and June 30 , 1990 , the civil penalty is \$100 if the order was issued between July 1 , 1990 and October 31 , 1996 , the civil penalty is \$200 if the order was issued between November 1 , 1996 and October 31 , 2006 , the civil penalty is \$300 "}\\
\midrule
{\bf Source}         &  {\bf Response}\\
Reference            & "the civil fine is \$ 200"\\
\emph{RAGbase}  & "the fine is \$ 250"\\
\emph{CPII-NLP}  & "The civil fine is \$ 500"\\
\emph{GPTChat}  &  "If the fine was issued between July 1, 1990, and October 31, 1996, the amount of the civil penalty for an uninsured operation is \$300, and the amount for an uninsured accident is \$500."\\
\emph{GPTLama} & The civil penalty for orders issued between July 1, 1990 and October 31, 1996 is \$500.\\
\bottomrule
\end{tabularx}
\caption{Examples of model generated responses compared to hand-written reference response from corpus.}
\label{tab:ExamplesOfGeneratedResponses}
\end{table}

We used the sandbox of the Amazon Mturk interface to set up the evaluation interface.  With 25 examples and four variants
 of the agent responses, the experiment included 100 unique human intelligence tasks (HITs).
Each HIT was assigned to three annotators and a total of nine raters participated in the experiment. 
Given the complexity of the task, all recruited annotators were colleagues from our lab (including the authors) and
were not paid for this task.


\section{Conclusion}

This paper presented a study of document-grounded response generation methods for information-seeking dialogue modeling
particularly including two methods utilizing one of the \emph{ChatGPT} large language models for this task. 
The comparison was conducted by both objective metrics as well as by human evaluation.
Objective evaluation results showed that
typical word-overlap based metrics are not suitable to fully assess the performance of these methods and human evaluation indicated that 
\emph{ChatGPT}-based models have strong potential in this domain, even exceeding appropriateness-scores for the human-authored reference responses.
Just providing dialogue context and a system-prompt specifying the domain and the role of the response generator
was sufficient to outperform human generated responses on the subjective appropriateness metric.
While the system which additionally utilized document-retrieved information showed the
highest appropriateness score, it was only marginally better than the system without additional retrieval step. 
However, the potential to access specific external information (or private information) not seen by the LLM during training
is essential in specific domains. Especially when information from external documents can be reliably retrieved and
methodically inserted into the LLM prompts can help reduce hallucinations.
However, human evaluation also visualized the challenges in assessing 
the accuracy and veracity of \emph{ChatGPT}-generated responses which can simultaneously appear very well formulated but factually wrong. 
In future work we plan to apply fact verification methods for assessing reliability of generated responses.

\bibliography{custom}

\newpage 

\appendix

\section{Appendix}

\subsection{\emph{GPTChat}: Example of input format}
\label{GPTChat_Exampleofinput}

\begin{table}[H]
\center
\scriptsize
\begin{tabularx}{\linewidth}{lX}
\toprule
\{"role":   "system",  "content":  & "Hello, welcome to the Department of Motor Vehicle information service agent - how can I help you?"\},\\
\{"role":   "user",     "content":   & "Hello, I forgot to update my address, can you help me with that?"\},\\
\{"role":   "assistant",  "content":  & "hi, you have to report any change of address to DMV within 10 days after moving. You should ... vehicles."\},\\
\{"role":   "user",      "content":  & "Can I do my DMV transactions online?"\}\\
\bottomrule
\end{tabularx}
\caption{Example of ChatCompletion input format of dialogues.}
\label{tab:ChatCompletionInptFormatExample}
\end{table}

\subsection{\emph{GPTChat}: Initial system prompts for each domain}
\label{ssec:InitialSysPromptsChatCompletion}

{\small
\begin{itemize}\itemsep-0.2em
\item[DMV]{"Hello, welcome to the Department of Motor Vehicle information service agent - how can I help you?"}
\item[SSA]{"Hello, I'm the service agent from the US Government Social Security Administration department and can help you with any questions about retirement, disability benefits, how to get or replace your Social Security card, and more.''}
\item[STA]{"Hi, I'm the service agent for the U.S. department for Federal Student Aid, which offers grants, loans, work-study, and more to help you pay for college or career school. I can answer your questions related to the Free Application for Federal Student Aid (FAFSA), types of student aid and the many ways to get help paying for college or career school."}
\item[VA]{"Hello, I'm the service agent for the U.S. Department of Veterans Affairs (VA) where service members, veterans, and their beneficiaries can apply for benefits services. I'm also linked with the Federal Benefits Unit (FBU) and can answer your questions about our benefits and services."}
\end{itemize}
}


\subsection{\emph{GPTLama}: Initial domain-specific prompts}
\label{ssec:InitialSysPrompts4GPTLama}

{\small
\begin{itemize}\itemsep-0.2em
\item[DMV]{"Question for the U.S. department of Motor Vehicles (DMV) service agent: "}
\item[SSA]{"Question for the US Government Social Security Administration (SSA) service agent: ''}
\item[STA]{"Question for the U.S. department for Federal Student Aid service agent: "}
\item[VA]{"Question for the U.S. department for Veterans Affairs (VA) service agent: "}
\end{itemize}
}

\subsection{\emph{GPTLama}: Domain-specific prompts before dialogue context}
\label{ssec:PreDialContextPrompts4GPTLama}
{\small
\begin{itemize}
\item[DMV]{"<system> Hello, I'm the service agent from the US Department of Motor Vehicles (DMV) Here is the conversation I had with the user before I received the question to be answered by you:</system>."}
\item[SSA]{<system> "Hello, I'm the service agent from the US Government Social Security Administration (SSA) department and can help you with any questions about retirement, disability benefits, how to get or replace your Social Security card, and more. Here is ... </system>.''}
\item[STA]{"<system> Hi, I'm the service agent for the U.S. department for Federal Student Aid, which offers grants, loans, work-study, and more to help you pay for college or career school. I can answer your questions related to the Free Application for Federal Student Aid (FAFSA), types of student aid and the many ways to get help paying for college or career school. Here is ...</system>."}
\item[VA]{"Hello, I'm the service agent for the U.S. Department of Veterans Affairs (VA) where service members, veterans, and their beneficiaries can apply for benefits services. I'm also linked with the Federal Benefits Unit (FBU) and can answer your questions about our benefits and services. Here is ...</system>."}
\end{itemize}
}

\subsection{\emph{GPTLama}: Example of prompt}
\label{ssec:ExamplePrompt4GPTLama}

{\small
query\_str = "Question for the US Department of Motor Vehicles (DMV) service agent: <user> My driver license has been suspended and I need help fixing this."

QA\_PROMPT\_TMPL = ("We have provided context information below: \textbackslash{n}\textbackslash" "---\textbackslash{n}"\textbackslash "<system> Hello, I'm the service agent from the US Department of Motor Vehicles (DMV)\textbackslash Here is the conversation I had with the user before I received the question to be answered by you:</system>. \textbackslash{n}\textbackslash" "And here is some context information I've extracted from my database:" "\{context\_str\}" "\textbackslash{n}---\textbackslash{n}" "Given this information, please answer the question as service agent from the US Department of Motor Vehicles: \{query\_str\}\textbackslash{n}")
}

\end{document}